\documentclass[letterpaper]{article}
\usepackage[preprint]{aaai2027}
\usepackage[hyphens]{url}
\usepackage{graphicx}
\usepackage{natbib}
\usepackage{caption}
\usepackage{amsmath,amssymb}
\usepackage{booktabs}
\usepackage{multirow}
\usepackage{tabularx}
\usepackage{array}

\newcommand{\systemname}{mmMind}
\title{Teaching Foundation Models to Read mmWave: Pose-Guided Kinematic Representation for Human Behavior Understanding}
\author{
    Duo Zhang\textsuperscript{\rm 1}\equalcontrib,
    Zhehui Yin\textsuperscript{\rm 1}\equalcontrib,
    Zhiyun Yao\textsuperscript{\rm 1},
    Haotong Qin\textsuperscript{\rm 2},
    Xusheng Zhang\textsuperscript{\rm 1},\\
    Hongliu Yang\textsuperscript{\rm 1},
    Jianyu Sun\textsuperscript{\rm 1},
    Junzhe Wang\textsuperscript{\rm 1},
    Zizhou Fan\textsuperscript{\rm 1},
    Michele Magno\textsuperscript{\rm 2},
    Daqing Zhang\textsuperscript{\rm 1,\rm 3}\corresponding
}
\affiliations{
    \footnotesize
    \textsuperscript{\rm 1}Peking University \quad
    \textsuperscript{\rm 2}ETH Zurich \quad
    \textsuperscript{\rm 3}Institut Polytechnique de Paris\\
    \{zhangduo,yinzhehui,zhiyunyao\}@stu.pku.edu.cn \quad
    haotong.qin@pbl.ee.ethz.ch \quad
    \{zhangxusheng,Hongliu\_yang\}@stu.pku.edu.cn \quad
    jys62544@gmail.com \quad
    \{wjz020606,zzfan25\}@stu.pku.edu.cn \quad
    michele.magno@pbl.ee.ethz.ch \quad
    dqzhang@sei.pku.edu.cn
}
\begin{document}

\maketitle

\begin{abstract}
Large language model agents need to perceive human behavior in physical environments.
Millimeter-wave (mmWave) radar provides a privacy-friendly and contactless sensing modality, but radar observations are difficult to align with language. Existing radar-language methods often rely on synthetic data or lack explicit supervision for human body structure and motion.
We present mmMind, a radar-language model that uses synchronized 3D pose as training-only supervision. A spatio-temporal radar encoder is pretrained to capture body configuration and motion dynamics, after which the pose head is removed so that inference requires radar alone. The learned radar representations are then aligned with an LLM for behavior captioning and spatio-temporal question answering. We also introduce mmMind-Bench, a real-world mmWave-language benchmark containing 17.9 hours of recordings from 23 participants across seven indoor environments. Experiments on captioning, question answering, and unseen-action generalization show that mmMind consistently outperforms existing radar-language baselines, while ablations confirm the importance of pose-guided pretraining.
\end{abstract}

\noindent
\textbf{WebPage:}
\url{https://duo-zhang-sensing.github.io/mmMind/}

\noindent
\textbf{Demo:}
\url{https://www.youtube.com/watch?v=ccmq6U8t-y0}

\section{Introduction}
\label{sec:intro}

Large language model (LLM) agents increasingly ground their responses in multimodal context, including text, images, audio, and video. 
Agents operating in everyday environments, however, must also perceive human behavioral context: what people are doing, and how their activities unfold \cite{fu2026deepsensemoe}. 
Access to such context is essential for agents to provide timely, behavior-aware assistance rather than relying solely on explicit user input.

Continuously acquiring such context remains challenging. 
Cameras raise privacy concerns in spaces such as bedrooms and bathrooms, while wearables depend on sustained user compliance~\cite{healthllm,video1}. 
Millimeter-wave (mmWave) radar offers a privacy-friendly, contactless, and low-cost alternative by sensing human from radio reflections without capturing recognizable imagery or requiring on-body devices. 
It has enabled applications including activity recognition, pose estimation, and physiological monitoring~\cite{smarthome,zhang2026mupose,liu2026mmjepa,ltfall}.

\begin{figure*}[t]
    \centering
    \includegraphics[width=\textwidth]{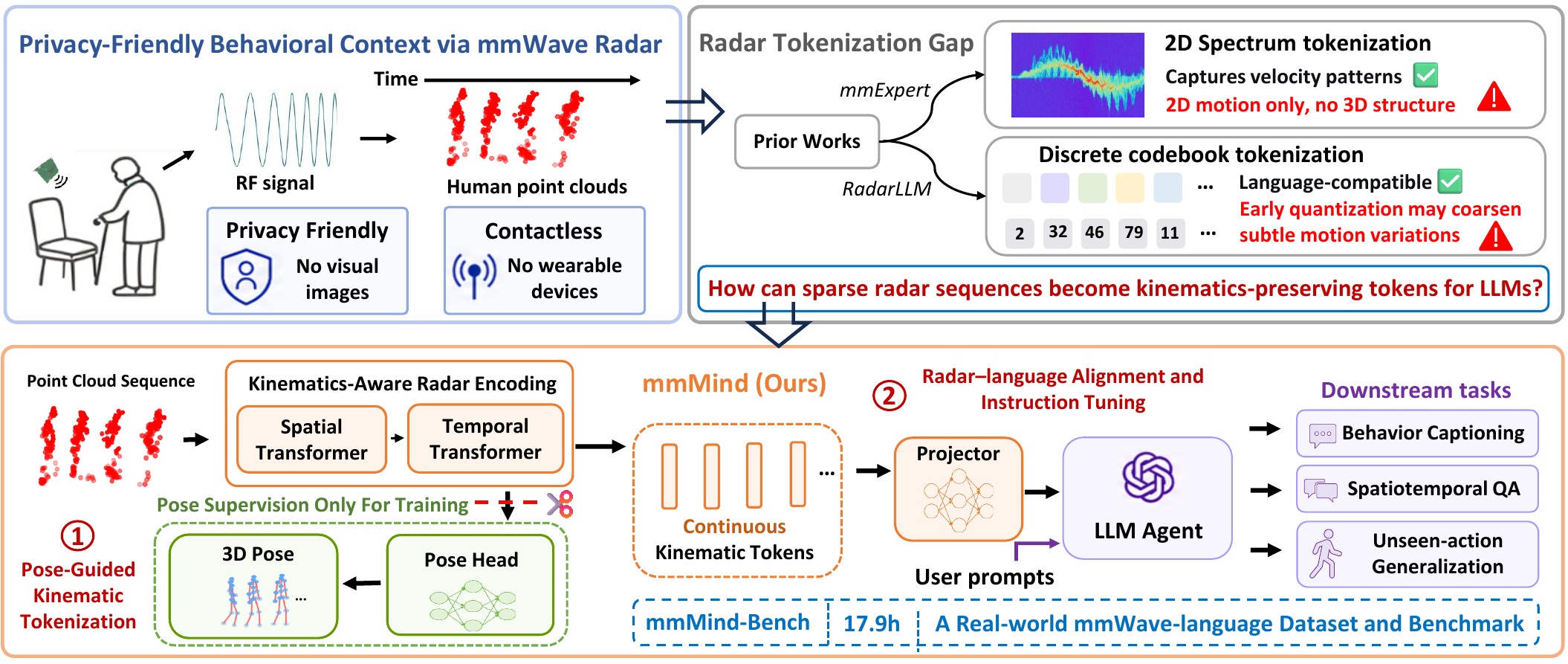}
\caption{
Overview of \systemname{}.
Existing radar-language approaches rely on 2D spectrum representations or
discrete codebook tokens, which may lose body kinematics or fine-grained
motion variations.
\systemname{} addresses these challenges through pose-guided kinematic
tokenization and mmMind-Bench, a real-world radar-language benchmark for
alignment and instruction tuning.
At inference, radar alone enables behavior captioning, spatio-temporal QA,
and unseen-action generalization.
}
    \label{fig:overview}
\end{figure*}

Yet mmWave is not a modality that an LLM can directly read.
Among common radar representations, temporal point clouds provide a
promising basis for behavior understanding because they retain the 3D
locations and radial velocities of body reflections
~\cite{Radhar,Pantomime,ptr,mmgpe}.
However, they remain sparse, noisy, and temporally irregular, with
behavioral semantics emerging from subtle changes in body configuration
and motion over time.
This raises a fundamental question:

\begin{quote}
\emph{How can sparse radar sequences be transformed into kinematically grounded representations and aligned with an LLM for behavior understanding?}
\end{quote}

As illustrated in Fig.~\ref{fig:overview}, answering this question requires
addressing two coupled challenges: preserving human kinematics in the radar
representation, and aligning that representation with language under real
sensing conditions.

\textbf{Challenge 1: Kinematic representation.}
Conventional mmWave systems map an entire sequence to a closed-set or
open-vocabulary label~\cite{Radhar,mmCLIP}, without describing how the
behavior unfolds.
Recent radar-language models instead construct interfaces between radar
observations and LLMs.
For example, mmExpert aligns 2D micro-Doppler spectra with language
\cite{mmExpert}, capturing velocity patterns but providing limited access to
explicit 3D body structure.
RadarLLM converts temporal point clouds into discrete motion codes using an
Aggregate VQ-VAE~\cite{RadarLLM}.
These approaches establish useful radar-language interfaces, but their
representation-learning objectives do not explicitly encourage the latent
space to preserve body configuration, global displacement, and local joint
dynamics.
Generic point-cloud LLMs are also not directly suited to this setting because
they are designed primarily for dense and largely static objects or scenes
\cite{pointmllm,hong20233dllm}.

To address this challenge, we propose
\emph{pose-guided kinematic representation learning}.
A 3D pose sequence provides a structured description of human kinematics, capturing both body configuration and its temporal evolution---the very information radar tokens should preserve.
We therefore pretrain a spatio-temporal radar encoder to recover synchronized poses, shaping its latent space toward kinematic structure. After pretraining, the pose head is removed, and the encoder produces continuous kinematic tokens from radar alone.

\textbf{Challenge 2: Real-world radar-language alignment.}
These tokens must then be mapped into the LLM embedding space and grounded in language through cross-modal alignment and instruction tuning.
This requires paired radar-language data with fine-grained behavior
annotations.
Synthetic data provide an elegant and scalable solution when such
annotations are scarce and have enabled important progress in prior work ~\cite{RadarLLM,mmExpert,mmCLIP,chen2023rf,gong2025data}.
Nevertheless, simulation alone cannot fully capture the sparsity,
multipath, clutter, and hardware noise of physical deployments, while
large-scale real-world language supervision remains limited.

We therefore introduce \textbf{mmMind-Bench}, to our knowledge the first
large-scale real-world mmWave-language dataset and benchmark.
It contains 17.9 hours of recordings from 23 participants, covering 27 daily-life and exercise-related behaviors.
Beyond action labels, mmMind-Bench provides fine-grained language annotations describing body states, motion trajectories, temporal
evolution, and behavior context, together with synchronized 3D poses,
spatio-temporal QA, and multi-turn dialogues.
These real-world annotations enable radar--language alignment through projector training and LLM instruction tuning, while supporting unified evaluation on captioning, question answering, interaction, and unseen-action generalization.

Building on these two components, we develop \textbf{\systemname{}} on
commercial off-the-shelf mmWave hardware.
\systemname{} achieves a BERTScore of 86.8 on radar captioning, 89.2\%
accuracy on spatio-temporal QA, and 91.2\% accuracy on unseen actions,
consistently outperforming existing radar--language baselines.
Ablations further validate the proposed design.
Our contributions are threefold:
\begin{itemize}
    \item We redefine radar tokenization as learning \textbf{kinematically grounded continuous representations} from training-only 3D pose supervision, enabling radar tokens to preserve body structure and motion evolution.

    \item We introduce \textbf{mmMind-Bench}, a large-scale real-world mmWave-language dataset and benchmark. The dataset and code will be released upon publication.

    \item We build, to our knowledge, the first mmWave-enabled foundation model that supports multi-turn reasoning and interaction for human
behavior understanding, achieving best performance across all
three benchmark tasks.
\end{itemize}

\section{Related Work}
\label{sec:related}
\subsection{Wireless Sensing and mmWave-Based Human Behavior Understanding}

Wireless sensing provides a privacy-friendly and contactless approach to
understanding human behaviors in physical environments \cite{wang2025wicg,li2025rethinking,wang2025freebfi,yao2025wicaliper,fan2026m4human}. 
Among various wireless sensing modalities, millimeter-wave (mmWave) radar is particularly promising due to its fine-grained spatial and motion sensing capability, supporting activity recognition~\cite{Radhar,yang2026bridging,zhang2026towards,waffle,zhang2025mmrotation,yang2026omnipc}, 
3D pose estimation~\cite{zhang2026mupose,fan2026mmpred,liang2026wave2body,liang2026wicompass,xue2021mmmesh,cao2025occmesh,zheng2026person}, 
and physiological monitoring~\cite{zhang2025breaking,zhang2024from,chang2024mmecare}. 
Most existing methods formulate behavior understanding as classification or regression over predefined task spaces. 
Contrastive approaches such as mmCLIP~\cite{mmCLIP} extend recognition to open-vocabulary categories, 
but still compress an entire behavior sequence into an isolated prediction.

\subsection{Multimodal Tokenization and Language Alignment}

Understanding-oriented multimodal LLMs commonly preserve continuous
features from modality-specific encoders and align them with the LLM
embedding space through projectors or query-based connectors
~\cite{llava,llava15}, with recent work further improving their
efficiency through structure-preserving token compression
~\cite{feng2026kd,zhang2026hcc}.
This paradigm has also been extended to
3D point clouds~\cite{pointmllm,hong20233dllm}. Discrete VQ-based
tokenization remains useful for compression and multimodal
generation~\cite{van2017neural}, but early quantization may impose an information
bottleneck before language alignment. 
Moreover, reconstruction-oriented objectives may not preserve the
task-relevant structures required for language understanding.

\subsection{Radar-Language Models and Benchmarks}

Recent studies connect mmWave sensing with language models through different interfaces. 
mmWave-QA textualizes radar observations for training-free reasoning with off-the-shelf LLMs~\cite{shin2026can}; mmExpert uses 2D micro-Doppler spectrograms~\cite{mmExpert}; and RadarLLM quantizes temporal point clouds into discrete motion codes through an Aggregate VQ-VAE~\cite{RadarLLM}. 
These interfaces may limit the preservation of fine-grained 3D body structure and motion dynamics. 
Moreover, trainable radar-language models rely heavily on synthetic data and focus mainly on radar-to-text generation. 
In contrast, \systemname{} learns pose-guided continuous radar tokens and is evaluated on captioning, spatio-temporal question answering, and unseen-action generalization using mmMind-Bench, a large-scale real-world radar-language benchmark.

\section{mmMind-Bench}
\label{sec:benchmark}

We introduce mmMind-Bench, a real-world benchmark that pairs synchronized mmWave sensing data with bilingual language annotations at multiple levels for human behavior understanding.
Data collection was conducted under IRB
approval with informed consent from all participants.

\subsection{Real-World Data Collection}
mmMind-Bench contains 17.9 hours of real-world recordings from 23 participants
aged 21--62 across seven indoor environments. The dataset covers diverse
daily activities (e.g., walking, sitting, bending, computer usage, lying
down, sweeping, jogging, falling, drinking, and phone usage), exercise
motions (e.g., sit-ups, squats, jumping jacks, rope skipping, and boxing),
and upper-body gestures and interactions (e.g., pushing, pulling, drawing
circles, and vertical hand motions). The environments include diverse indoor
layouts and sensing conditions, such as living rooms, meeting rooms,
offices, and kitchens.

We retain raw radar measurements and derived point clouds from commercial 
60-GHz and 77-GHz devices, together with synchronized RGB-D videos and 3D
pose ground truth. All radar observations are transformed into a unified
extrinsic coordinate system, where the ground plane defines the \(x\)-\(y\)
plane for consistent spatial representation. 
To improve the quality of radar point clouds, we adopt our previous signal
processing pipeline~\cite{cmcfar,zhang2026mupose}, which produces denser
point clouds with reduced multipath interference and provides richer
spatio-temporal cues for human motion understanding. 
The use of multiple frequency bands and environments introduces variations
in reflection characteristics, clutter, and hardware conditions. RGB-D video
is used only for language annotation, while 3D pose provides training-only
kinematic supervision for the radar encoder. 
Details of signal processing
and point-cloud generation are provided in the appendix.

\begin{figure}[t]
    \centering
    \includegraphics[width=\columnwidth]{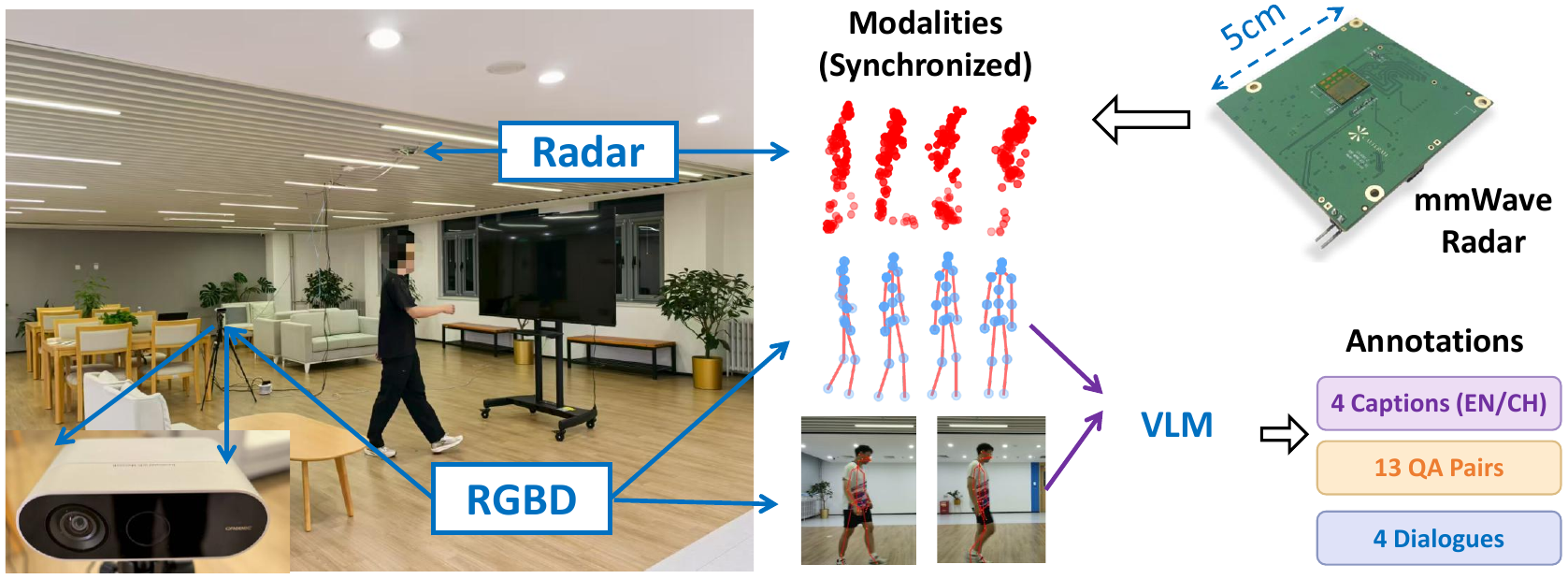}
    \caption{Overview of mmMind-Bench data collection and annotation. }
    \label{fig:mmMind_benchmark}
\end{figure}

\subsection{Radar-Grounded Language Annotation}

We segment each continuous recording into 5-second clips, yielding 75 radar frames per clip.

We adopt a human-seeded, pose-aware VLM pipeline.
Human-written examples define the annotation prompt.
Each synchronized 3D pose sequence is converted by rule-based processing
into a structured JSON summary of body states, posture transitions, limb
motion, direction, trajectory, speed, and event order.
InternVL3 \cite{zhu2025internvl3} then uses the synchronized RGB-D clip and the JSON
summary to draft captions, QA pairs, and dialogues.

The prompt restricts annotations to radar-observable behavior and excludes
appearance, identity, clothing, fine-grained object identity, and other
visual-only information. Automatic checks enforce format, bilingual
completeness, and the absence of prohibited attributes. Importantly, all
generated annotations are manually reviewed and revised for factual
correctness, kinematic consistency, radar observability, ambiguity, and
English--Chinese quality before inclusion in the benchmark. Only these
human-verified annotations are used for training and evaluation.

Each clip contains concise and detailed captions in English and Chinese,
13 single-turn QA pairs, and four multi-turn dialogues, covering behavior,
body state, limb motion, spatial relations, trajectory, speed, temporal
ordering, and cross-turn reasoning. 

The 5-second window standardizes the benchmark rather than constraining the
model, which supports variable-length inputs. A long-horizon extension based on temporal memory and query-driven radar
re-perception is described in Appendix~\ref{sec:supp_long_term_agent}.

\subsection{Benchmark Tasks and Splits}

mmMind-Bench evaluates three capabilities:
\textbf{radar captioning} generates fine-grained behavior descriptions;
\textbf{spatio-temporal question answering} covers single-turn and
contextual multi-turn reasoning over body states, spatial relations,
trajectories, motion changes, and event ordering; and
\textbf{unseen-action recognition} measures zero-shot generalization to
actions absent from training.

We construct participant-, environment-, and session-disjoint train,
validation, and test sets containing approximately 80\%/10\%/10\% of the
clips. Participants and environments are mutually exclusive across splits,
ensuring unseen individuals and indoor environments at evaluation. To avoid
leakage from the 5-second window with a 1-second stride, complete recording
sessions are partitioned before windowing, so overlapping or adjacent clips
never cross split boundaries; all annotations for the same clip follow the
same split.
For unseen-action evaluation, we construct action-disjoint test sets
containing multiple held-out behavior categories that are excluded from
training. 
All clips and associated pose and language annotations from the
held-out categories are excluded from radar encoder pretraining, projector
alignment, instruction tuning, and hyperparameter selection.

\section{mmMind}
\label{sec:method}

\systemname{} learns continuous radar tokens that preserve human
kinematics through training-only 3D pose supervision, and aligns them with
an LLM for behavior understanding.

\subsection{Overview and Problem Formulation}

Let a radar sequence be
\begin{equation}
    \mathcal{X}
    =
    \{\mathbf{X}_1,\mathbf{X}_2,\ldots,\mathbf{X}_T\},
\end{equation}
where each frame
\(\mathbf{X}_t \in \mathbb{R}^{N_t \times 4}\)
contains \(N_t\) radar points, each represented as
\((x,y,z,v)\), including its 3D location and radial velocity.

During encoder pretraining, the radar sequence is paired with a
synchronized 3D pose sequence
\begin{equation}
    \mathcal{S}
    =
    \{\mathbf{S}_1,\mathbf{S}_2,\ldots,\mathbf{S}_T\},
\end{equation}
where each \(\mathbf{S}_t\) contains the 3D coordinates of \(J\) body
joints. Pose is used only as training supervision and is unavailable at
inference time.

Given a radar sequence \(\mathcal{X}\) and a textual instruction
\(\mathbf{q}\), the goal is to generate a language response
\(\mathbf{a}=(a_1,\ldots,a_L)\):
\begin{equation}
    p(\mathbf{a}\mid\mathcal{X},\mathbf{q}).
\end{equation}

\begin{figure*}[t]
    \centering
    \includegraphics[width=\textwidth]{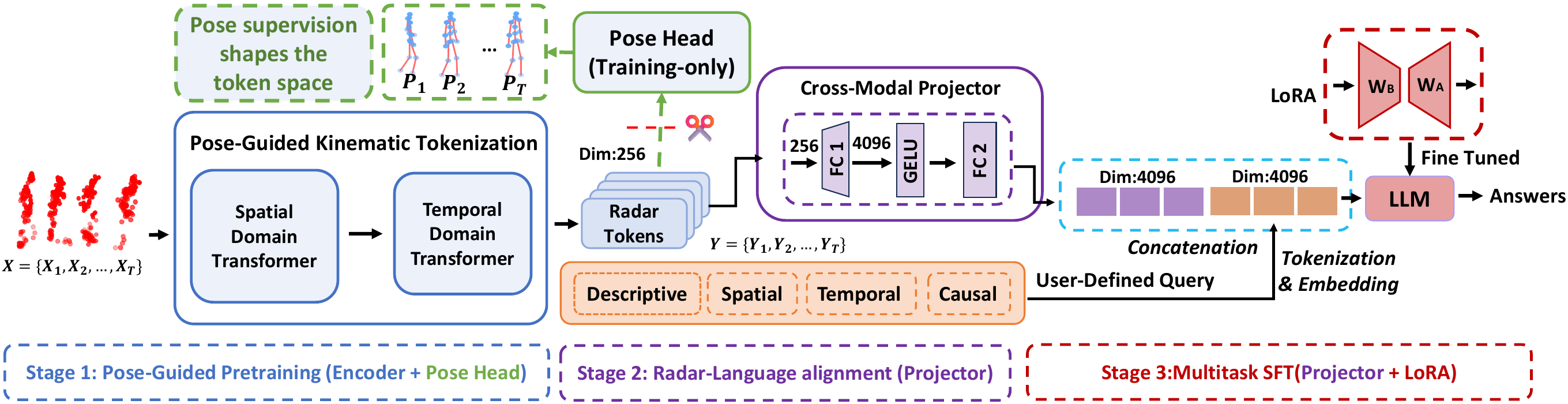}
\caption{
The detailed structure of the proposed \systemname{} model.
Pose supervision shapes continuous radar tokens during pretraining, which are aligned with an LLM for behavior understanding through captioning, question answering, and dialogue.}
\label{fig:mmmind_architecture}
\end{figure*}

As shown in Fig.~\ref{fig:mmmind_architecture}, a spatial encoder first
aggregates each variable-sized radar frame, and a temporal Transformer
models the evolution of the resulting features. During pose-guided
pretraining, a lightweight pose head encourages these representations to
preserve body configuration and motion dynamics. The pose head is removed
after pretraining, leaving a radar encoder that produces continuous
kinematic tokens. A cross-modal projector maps these tokens into the LLM
embedding space, where they are combined with the textual instruction for
language generation.

Training consists of three stages. First, the radar encoder is pretrained
with synchronized 3D pose supervision. Second, the encoder and LLM are
frozen while the projector learns radar-language alignment. Finally, the
encoder remains frozen, while the projector and LLM LoRA adapters are
instruction-tuned on captioning, single-turn QA, and multi-turn dialogue
data.

\subsection{Pose-Guided Kinematic Tokenization}

A single radar frame provides only a sparse and partial observation of the
human body, as reflected points vary with body orientation, self-occlusion,
and limb motion. However, human motion is temporally continuous, and body
parts missing in one frame may become observable in neighboring frames.
We therefore jointly model spatial point relationships and temporal
motion continuity.

\paragraph{Input normalization and augmentation.}
Each frame is standardized to \(M=150\) points by random subsampling or
zero-padding with a validity mask.  We normalize the \(x\)- and
\(y\)-coordinates relative to the first-frame horizontal centroid, and clip
radial velocity to \([-1,1]\). Although radial velocity captures only
line-of-sight motion, it remains useful for distinguishing moving limbs from
the torso.
During training, we apply
synchronized scaling, yaw rotation, and horizontal translation to the radar
and pose sequences, and add coordinate jitter to the radar points to improve
robustness to geometric and measurement variations.

\paragraph{Spatio-temporal radar encoder.}
Each four-dimensional point \((x,y,z,v)\) is projected into a
\(d_r\)-dimensional feature, while its 3D coordinates are separately mapped
to a learned spatial positional embedding. 
Let \(\widetilde{\mathbf{X}}_t\) denote the standardized point set and
\(\mathbf{m}_t\) its padding mask.
A spatial Transformer models
relationships among valid points, followed by learnable-query pooling that
aggregates each frame into one representation:
\begin{equation}
    \mathbf{h}_t =
    \operatorname{Pool}\!\left(
    E_{\mathrm{spa}}(\widetilde{\mathbf{X}}_t,\mathbf{m}_t)
    \right).
\end{equation}
A temporal Transformer with rotary position embeddings then integrates
information across frames:
\begin{equation}
    \mathbf{Y}
    =
    E_{\mathrm{tmp}}(\mathbf{h}_1,\ldots,\mathbf{h}_T)
    =
    \{\mathbf{y}_1,\ldots,\mathbf{y}_T\}.
\end{equation}
Spatial attention captures the body structure implied by simultaneous
reflections, whereas temporal attention exploits motion continuity and
complementary observations across time. Each \(\mathbf{y}_t\) forms one
continuous kinematic token for the corresponding radar frame.

\paragraph{Pose-guided pretraining.}
A lightweight pose head maps each radar token to the 3D coordinates of
\(J=17\) body joints:
\begin{equation}
\widehat{\mathbf{S}}_t=D_{\mathrm{pose}}(\mathbf{y}_t).
\end{equation}
We optimize the radar encoder with a kinematic objective:
\begin{equation}
\mathcal{L}_{\mathrm{kin}}
=
\lambda_p\mathcal{L}_{p}
+\lambda_r\mathcal{L}_{r}
+\lambda_l\mathcal{L}_{l}
+\lambda_b\mathcal{L}_{b}.
\end{equation}
The four terms supervise joint positions, global root motion, local joint
articulation, and skeletal consistency, respectively.
Specifically, \(\mathcal{L}_{p}\) adopts a joint-weighted \(L_1\) loss that
emphasizes distal joints, while the velocity losses capture body
translation and root-relative articulation. \(\mathcal{L}_{b}\) preserves
skeletal proportions.
After pretraining, the pose head is removed, and the radar encoder
produces continuous kinematic tokens from radar observations alone.

\subsection{Continuous Radar-Language Alignment}

After pose-guided pretraining, the radar encoder produces a sequence of
continuous kinematic tokens
\(\mathbf{Y}=\{\mathbf{y}_1,\ldots,\mathbf{y}_T\}\).
We introduce a lightweight projector \(g_{\phi}\) to map each radar token
from the encoder dimension \(d_r\) into the LLM embedding space:
\begin{equation}
    \mathbf{z}_t = g_{\phi}(\mathbf{y}_t),
    \qquad
    \mathbf{Z}=\{\mathbf{z}_1,\ldots,\mathbf{z}_T\}.
\end{equation}
The projector consists of two linear layers with normalization and GELU
activation. Each projected token replaces a frame-indexed placeholder in
the input sequence, preserving the one-to-one temporal correspondence
between radar frames and continuous LLM embeddings.

We then perform an initial radar-language alignment stage using
caption-only supervision. The training data contain concise and
attribute-level behavior descriptions in both English and Chinese, but
exclude question answering and multi-turn dialogue. Given the projected
radar tokens and a captioning instruction, the model is optimized with the
autoregressive language-modeling objective: 
\begin{equation}
\mathcal{L}_{\mathrm{align}}
=
-\sum_{\ell=1}^{L}
\log p(c_\ell \mid Z,q,c_{<\ell}).
\end{equation}
where \(c_\ell\) is the \(\ell\)-th token of the caption, and \(L\) is the caption length.
Only assistant-response tokens contribute to the loss. The radar encoder
and LLM remain frozen, and only the projector is updated. This stage first
establishes a stable correspondence between radar kinematics and basic
behavioral language before the subsequent instruction tuning on more
complex reasoning tasks.

\subsection{Multitask Instruction Tuning}

After radar-language alignment, we instruction-tune
\systemname{} on behavior captioning, spatio-temporal question answering,
and multi-turn dialogue. For reasoning-oriented QA, we additionally provide
kinematic rationales that describe observable motion evidence, such as body
state transitions, trajectories, and temporal changes.

The projected radar tokens are concatenated with the textual context and
processed by the LLM. Let
\(\mathbf{s}=(s_1,\ldots,s_L)\) denote the complete textual sequence,
including system instructions, user queries, dialogue history, and assistant
responses. We optimize the masked autoregressive objective:
\begin{equation}
    \mathcal{L}_{\mathrm{SFT}}
    =
    -\sum_{\ell=1}^{L}
    m_{\ell}
    \log p
    \left(
        s_{\ell}
        \mid
        \mathbf{Z},s_{<\ell}
    \right),
\end{equation}
where \(m_{\ell}=1\) if \(s_{\ell}\) is an assistant-response token and
\(m_{\ell}=0\) otherwise. For multi-turn dialogues, assistant tokens from
all turns contribute to the loss, while system instructions, user messages,
radar-prefix tokens, and padding tokens are masked out.

During instruction tuning, the radar encoder remains frozen, while the
projector and LoRA adapters of the LLM are optimized \cite{hu2022lora}. This preserves the
pose-guided radar representation while adapting the LLM for language-based
behavior understanding.

\section{Experiments}
\label{sec:experiments}

\subsection{Experimental Setup}

\begin{table*}[t]
\centering
\small
\setlength{\tabcolsep}{3.8pt}
\renewcommand{\arraystretch}{1.08}

\caption{
Main results on mmMind-Bench.
Best and second-best results are shown in bold and underlined, respectively.
All trainable baselines are retrained on the same benchmark splits and
language annotations whenever applicable.
}
\label{tab:main_results}

\begin{tabular*}{\textwidth}{
@{\extracolsep{\fill}}
l
l
cc
cc
c
@{}
}
\toprule
\multirow{2}{*}{Method}
& \multirow{2}{*}{Representation}
& \multicolumn{2}{c}{Captioning}
& \multicolumn{2}{c}{Question Answering}
& \multirow{2}{*}{\shortstack{Unseen Action\\Acc. $\uparrow$}} \\
\cmidrule(lr){3-4}
\cmidrule(lr){5-6}
&
& METEOR $\uparrow$
& BERTScore $\uparrow$
& Single-turn $\uparrow$
& Multi-turn $\uparrow$
& \\
\midrule

\multicolumn{7}{@{}l}{\textit{General-purpose LLM/VLMs}} \\[1pt]

Qwen3-8B (Text Only)
& Instruction only
& 16.2
& 61.4
& 34.7
& 31.9
& 18.6 \\

mmWave-QA-style~\cite{shin2026can}
& Radar textualization
& 29.4
& 74.1
& 69.8
& 55.7
& 62.3 \\

Qwen3-VL-32B-Instruct~\cite{Qwen3-VL}
& Rendered PC video
& 34.6
& 67.9
& 69.5
& \underline{61.4}
& 76.7 \\

\midrule
\multicolumn{7}{@{}l}{\textit{Radar-specific models}} \\[1pt]

mmExpert~\cite{mmExpert}
& Micro-Doppler spectra
& 29.5
& 63.2
& 66.1
& --
& 67.4 \\

RadarLLM~\cite{RadarLLM}
& Discrete radar tokens
& \underline{36.2}
& \underline{75.2}
& \underline{74.6}
& --
& 74.8 \\

mmCLIP~\cite{mmCLIP}
& Contrastive radar-text
& --
& --
& --
& --
& \underline{79.7} \\

\midrule

\textbf{\systemname{} (Ours)}
& \textbf{Continuous tokens}
& \textbf{46.5}
& \textbf{86.8}
& \textbf{89.2}
& \textbf{84.7}
& \textbf{91.2} \\

\bottomrule
\end{tabular*}
\end{table*}

\paragraph{Tasks and splits.}
We evaluate \systemname{} on radar captioning, spatio-temporal question
answering, and unseen-action recognition using the participant-,
environment-, and session-disjoint splits in mmMind-Bench.
For unseen-action evaluation, the tested categories are excluded from both
radar encoder pretraining and instruction tuning. No method uses RGB-D
video or 3D pose at inference time.

\paragraph{Baselines.}
All trainable baselines are reimplemented and fully retrained on
mmMind-Bench; all reported results are obtained from our reproductions.
Whenever supported, they use the same captioning, QA, and dialogue
annotations as \systemname{}.
\textbf{Qwen3-8B (Text Only)} receives only the instruction and dialogue
history and is instruction-tuned on the same language annotations.
\textbf{mmWave-QA-style}~\cite{shin2026can} textualizes each point-cloud
sequence using spatial, trajectory, and radial-velocity statistics before
processing it with Qwen3-8B.
\textbf{Qwen3-VL-32B-Instruct}~\cite{Qwen3-VL} is fine-tuned on the same
language supervision using synchronized front- and side-view point-cloud
videos, with point color encoding radial velocity.
\textbf{mmExpert}~\cite{mmExpert} is retrained using micro-Doppler spectra
generated from the same raw radar recordings.
\textbf{RadarLLM}~\cite{RadarLLM} is retrained on the same 75-frame clips,
which are directly tokenized by its Aggregate VQ-VAE without temporal
resampling.
\textbf{mmCLIP}~\cite{mmCLIP} is retrained on the same splits for
unseen-action recognition.
A ``--'' indicates that the corresponding architecture does not support
the task.

\paragraph{Implementation Details.}
\systemname{} uses Qwen3-8B as the language backbone \cite{yang2025qwen3}.
The radar encoder outputs 256-dimensional frame tokens, followed by a
two-layer projector for radar--language alignment.
During instruction tuning, the radar encoder is frozen and only the
projector and LoRA adapters are optimized.
Training details and hyperparameter settings are provided in the
supplementary material.

\paragraph{Metrics.}
For radar captioning, we report METEOR, which measures lexical agreement
with stemming and synonym matching, and BERTScore-F1, which measures
contextual semantic similarity between generated and reference
descriptions. 
For QA, generated answers are lowercased, stripped of punctuation and
articles, and matched to a fixed canonical answer set using a predefined
synonym mapping; unmatched responses are counted as incorrect.
Single-turn accuracy is macro-averaged across question categories.
For multi-turn QA, turn-level accuracy is first computed within each
dialogue and then averaged across dialogues, with the complete preceding
dialogue history provided at every turn.
For unseen-action recognition, tested action categories are excluded from
all training stages. Generated action names are normalized to the held-out
label vocabulary before computing accuracy.
All metrics are reported on a 0--100 scale.

\subsection{Main Results on mmMind-Bench}

Table~\ref{tab:main_results} compares representative strategies for
connecting radar observations with language models.
\systemname{} achieves the best performance across all evaluated tasks.
The text-only baseline performs substantially worse, indicating that
language and dataset priors alone provide limited information about the
observed behavior.
Radar textualization and point-cloud video rendering expose additional
motion cues to general-purpose models, but introduce intermediate
abstraction or rendering steps.
Notably, the Qwen3-VL baseline uses a substantially larger 32B model and
rendered video inputs, whereas \systemname{} operates on compact continuous
radar tokens with an 8B language backbone.

\paragraph{Radar captioning.}
\systemname{} achieves a METEOR score of 46.5 and a BERTScore of 86.8,
outperforming the strongest baselines by 10.3 and 11.6 points,
respectively.
It consistently surpasses textualized radar descriptions, rendered
point-cloud videos, micro-Doppler spectra, and discrete radar tokens.
These results indicate that continuous kinematic tokens better preserve
body configuration and motion evolution for language generation.

\paragraph{Spatio-temporal question answering.}
\systemname{} obtains 89.2\% accuracy on single-turn QA, improving over the
strongest baseline by 14.6 points.
It further achieves 84.7\% accuracy on multi-turn dialogue, exceeding the
Qwen3-VL-32B baseline by 23.3 points.
These results show that the learned radar tokens support both direct
questions about observed motion and context-dependent follow-up questions
across dialogue turns.

\paragraph{Unseen-action recognition.}
On action categories excluded from both radar pretraining and instruction
tuning, \systemname{} achieves 91.2\% accuracy, outperforming the
contrastive mmCLIP baseline by 11.5 points.
The consistent improvement over textualized, rendered, micro-Doppler, and
discrete representations suggests that kinematically grounded continuous
tokens transfer more effectively to behaviors not observed during
training.

\subsection{Qualitative Results}
\label{sec:qualitative}

\begin{figure*}[t]
    \centering
    \includegraphics[width=\textwidth]{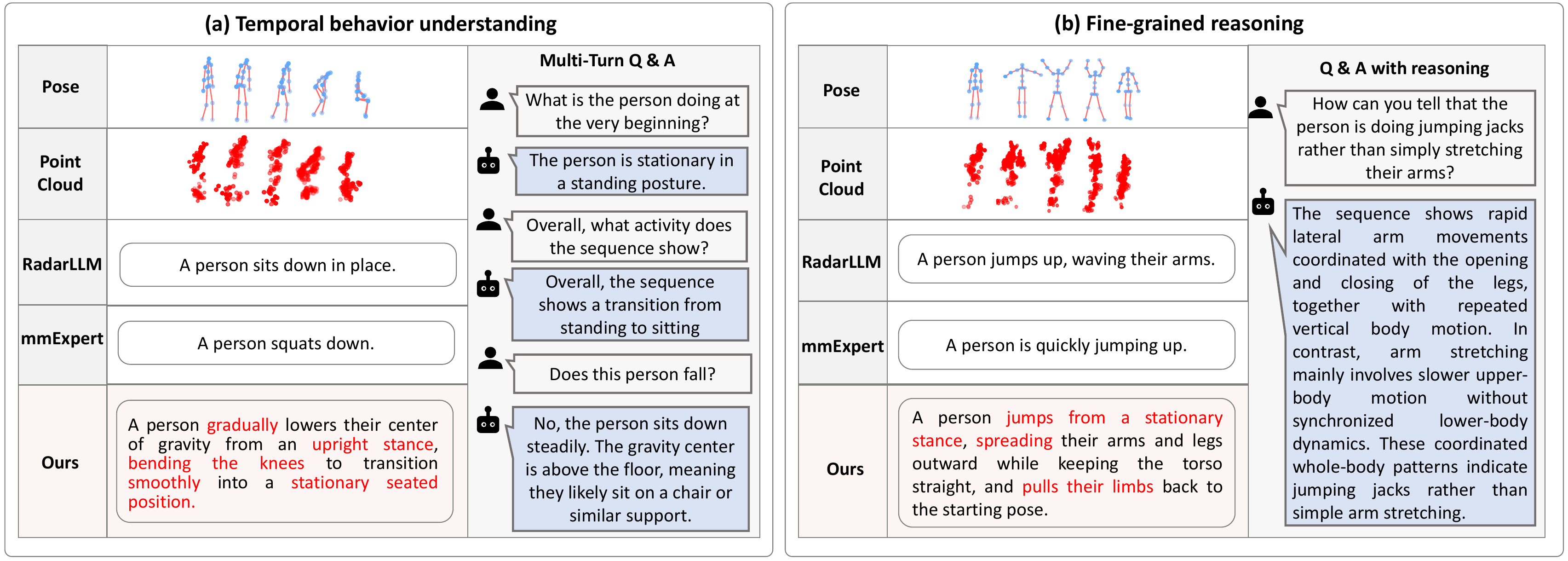}
    \caption{
    Qualitative results on mmMind-Bench.
    (a) \systemname{} captures the transition from standing to sitting and
    maintains consistent answers across dialogue turns.
    (b) It distinguishes jumping jacks from arm stretching using rapid arm
    motion, coordinated arm-leg movement, and repeated vertical motion.
    Synchronized 3D poses are shown only for visualization.
    }
    \label{fig:qualitative_mmMind}
\end{figure*}

Figure~\ref{fig:qualitative_mmMind} shows that \systemname{} captures more
fine-grained body states and motion evolution than RadarLLM and mmExpert.
In Fig.~\ref{fig:qualitative_mmMind}(a), the baselines recognize a general
downward or sitting motion, whereas \systemname{} describes the transition
from an upright stance to a stable seated posture, including the gradual
lowering of the body and knee flexion. Its responses also remain consistent
across dialogue turns, correctly identifying the initial state, summarizing
the overall transition, and distinguishing controlled sitting from falling.

In Fig.~\ref{fig:qualitative_mmMind}(b), all methods recognize a jumping
motion, but only \systemname{} describes the coordinated opening and closing
of both the arms and legs. When asked to distinguish jumping jacks from arm
stretching, it grounds the answer in rapid lateral arm motion, synchronized
lower-body dynamics, and repeated vertical body movement. These examples
illustrate that the learned radar tokens support both detailed behavior
description and context-dependent reasoning over temporal observations.

\subsection{Ablation Study}
\label{sec:ablation}

We ablate the key components of \systemname{} while keeping the language
backbone, data splits, and instruction-tuning protocol unchanged.
\textit{w/o Pose Pretraining} removes Stage-I pose-guided pretraining and
learns the radar encoder only through subsequent language supervision.
\textit{w/o Auxiliary Kinematic Losses} retains pose pretraining with the
joint-position loss, but removes the global-velocity, local joint-velocity,
and bone-length constraints.
\textit{w/o Temporal Encoder} removes the radar-side temporal Transformer
while retaining the ordered frame tokens provided to the LLM. 
\textit{w/ VQ Tokens} replaces the continuous frame-level representations
with discrete codes learned by a VQ-VAE.

\begin{table}[ht]
\centering
\footnotesize
\setlength{\tabcolsep}{3.0pt}
\renewcommand{\arraystretch}{1.08}
\caption{
Ablation results on mmMind-Bench.
Captioning is evaluated by BERTScore; the remaining columns report
accuracy. All metrics are on a 0--100 scale.
}
\label{tab:ablation}

\begin{tabular}{@{}lcccc@{}}
\toprule
& \multicolumn{1}{c}{Captioning}
& \multicolumn{2}{c}{Question Answering}
& \multicolumn{1}{c}{Unseen} \\
\cmidrule(lr){2-2}
\cmidrule(lr){3-4}
\cmidrule(lr){5-5}
Variant
& BERT $\uparrow$
& Single $\uparrow$
& Multi $\uparrow$
& Acc. $\uparrow$ \\
\midrule

No pose pretrain.
& 69.8
& 68.6
& 63.9
& 69.4 \\

No aux. kin. losses
& 84.1
& 85.0
& 79.1
& 86.3 \\

No temporal enc.
& 77.5
& 81.2
& 73.8
& 73.1 \\

VQ tokens
& 76.6
& 79.4
& 73.5
& 81.9 \\

\midrule

\textbf{Full model}
& \textbf{86.8}
& \textbf{89.2}
& \textbf{84.7}
& \textbf{91.2} \\

\bottomrule
\end{tabular}
\end{table}

As shown in Table~\ref{tab:ablation}, removing pose-guided pretraining
causes the largest degradation across all tasks, reducing captioning,
single-turn QA, multi-turn QA, and unseen-action accuracy by 17.0, 20.6,
20.8, and 21.8 points, respectively.
This result suggests that language supervision alone is insufficient to
organize sparse radar observations into effective behavior representations.

Removing the auxiliary kinematic losses produces smaller but consistent
drops of 2.7--5.6 points.
The larger reductions on multi-turn QA and unseen-action recognition
indicate that velocity and bone-length constraints complement joint-position
supervision by preserving motion dynamics and skeletal structure.
Removing the temporal encoder decreases multi-turn QA by 10.9 points and
unseen-action accuracy by 18.1 points, highlighting the importance of
cross-frame modeling for temporal reasoning and generalization.
Finally, replacing continuous representations with VQ tokens reduces
performance by 9.3--11.2 points across tasks, supporting the benefit of
avoiding early quantization when encoding fine-grained kinematic variations.

\section{Limitations and Conclusion}
\label{sec:conclusion}

\paragraph{Limitations and future work.}
\systemname{} remains less reliable for estimating precise quantities such
as speed, turning angle, and repetition count, especially under sparse
reflections or partial motion cycles. Future work will explore physics-aware
reinforcement learning with verifiable radar-derived rewards to improve
numerical accuracy, physical consistency, and uncertainty calibration.
Human-preference optimization may further improve the naturalness and
helpfulness of the model's reasoning and responses.

In multi-person scenarios, the current system relies on an external
clustering and tracking pipeline to separate the mixed radar point clouds
into person-specific sequences, after which each sequence is processed
independently by \systemname{}. Consequently, its performance depends on the
accuracy and temporal consistency of the preceding clustering and tracking
results, particularly when individuals are close to each other or interact.
Future work will investigate end-to-end multi-person radar point-cloud
understanding that jointly performs person separation, tracking, behavior
understanding, and interaction reasoning.

\paragraph{Conclusion.}
We presented \systemname{}, a radar-language agent that learns continuous
kinematic tokens from sparse temporal mmWave observations through
training-only pose supervision. We also introduced mmMind-Bench, a
large-scale real-world radar-language benchmark with synchronized 3D poses,
fine-grained behavior descriptions, spatio-temporal QA, and multi-turn
dialogues. Together, they demonstrate the potential of mmWave sensing as a
privacy-friendly perceptual interface for human-centered agents.

\section*{Acknowledgments}

This work was supported by the Beijing Natural Science Foundation
(International Scientists Project) under Grant No.~IS26037.
We thank Jiazun Chen of ByteDance Seed for valuable discussions on
multimodal large language models and related research directions.
We also thank all volunteers who assisted with data collection.
We acknowledge Yingbo Cloud for providing computational resources that
supported model training and evaluation.
We thank Peking University's New Yanyuan Campus for providing
experimental facilities and venue support for data collection.

\clearpage
\appendix
\setcounter{secnumdepth}{2}
\suppressfloats[t]
\begin{center}
    {\LARGE\bfseries Appendix}
\end{center}

This appendix provides additional details of \systemname{} and
mmMind-Bench. RGB-D videos and 3D poses are used only for language
annotation or training-time supervision; all reported inference results use
radar observations alone.

\begin{figure}[t]
    \centering
    \includegraphics[width=\columnwidth]{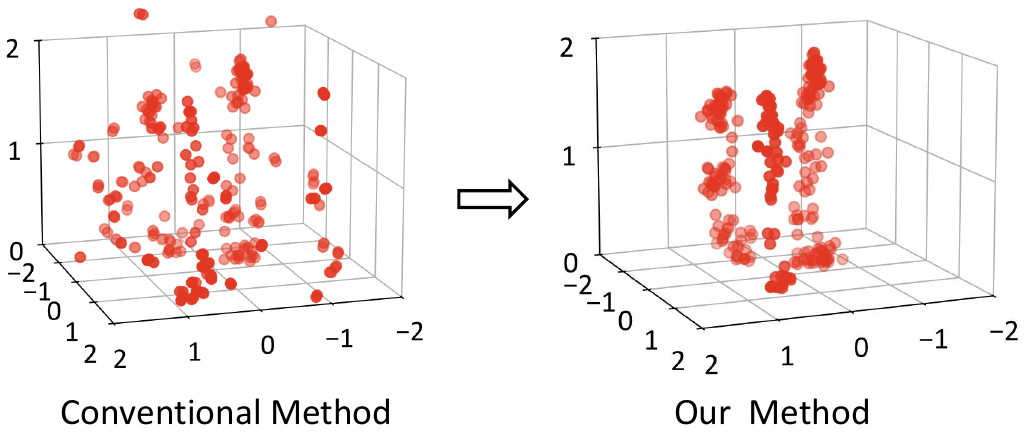}
    \caption{
    Comparison of human point clouds generated by conventional processing
    and the adopted pipeline. The latter recovers denser human reflections
    while reducing multipath-induced artifacts.
    }
    \label{fig:PointCloud}
\end{figure}

\section{mmWave Point Cloud Generation}
\label{sec:supp_point_generation}

mmWave point clouds provide the physical input to mmMind-Bench, and their
quality determines how much posture and motion information is available to
downstream models. Conventional radar processing often produces sparse human
reflections and multipath-induced ghost points in indoor environments. To
improve point-cloud quality, we adopt our prior processing methods:
ETCM-CFAR~\cite{cmcfar} for recovering weak human reflections and
AoA--AoD consistency checking~\cite{zhang2026mupose} for suppressing
multipath artifacts. As illustrated in Fig.~\ref{fig:PointCloud}, the
resulting point clouds contain denser and cleaner human reflections than
those produced by conventional processing.

\paragraph{Point-cloud formation.}
For each radar frame, raw measurements are organized as a radar cube over
fast time, slow time, and antenna channels. Range and Doppler FFTs are first
applied to obtain Range--Doppler representations, followed by target
detection and angle estimation. Each detected reflection is converted into a
4D radar point,

\begin{equation}
    \mathbf{p}_i=(x_i,y_i,z_i,v_i),
\end{equation}

where \((x_i,y_i,z_i)\) denotes its 3D location and \(v_i\) denotes its
radial velocity.

Using calibrated extrinsic parameters, radar points are transformed from the
sensor coordinate system into a unified room-level world coordinate system.
The ground plane is defined as the \(x\)-\(y\) plane and the vertical
direction as the \(z\)-axis. Consequently, \(x\) and \(y\) describe the
horizontal location of a reflection in the environment, while \(z\)
represents its absolute height. This calibration provides consistent spatial
representations across different radar placements and environments. The
transformed points are then used for subsequent processing and
radar-language modeling.

\paragraph{Why point clouds instead of dense radar representations.}
Dense radar heatmaps and high-dimensional radar cubes preserve rich
low-level signal responses, but they are highly redundant for human behavior
understanding. Human reflections occupy only a small portion of the sensing
volume, while most cells represent empty space, static background, clutter,
or noise. These dense representations therefore require substantial storage
and computation for information that is often unrelated to human motion.

Moreover, although radar cubes encode range, Doppler, azimuth, and elevation,
these physical properties remain implicitly distributed over a dense grid.
It is therefore difficult to directly associate individual reflections with
their 3D locations, velocities, and underlying kinematic structures.

In contrast, each radar point explicitly represents a physical reflection
with its 3D location and radial velocity. Temporal radar point clouds thus
provide a compact 4D representation that naturally aligns with human
kinematics and facilitates language-based behavior understanding. We
therefore adopt temporal radar point clouds as the primary representation in
mmMind-Bench.

\paragraph{Density enhancement with ETCM-CFAR.}
Conventional spatial CFAR estimates the noise level of a Range--Doppler bin
from neighboring reference cells. Although suitable for distant point
targets, this assumption can be problematic for short-range human sensing.
A human body is an extended target occupying multiple adjacent bins, and
strong torso reflections may enter the reference region and increase the
estimated noise level. As a result, weaker reflections from the arms, legs,
and other body parts can be suppressed by an excessively high detection
threshold, producing sparse and incomplete point clouds.

We address this issue using Extended Target Clutter Map CFAR
(ETCM-CFAR)~\cite{cmcfar}. Rather than estimating noise from spatially
adjacent cells, ETCM-CFAR maintains an independent temporal noise estimate
for each Range--Doppler bin. Given the current power \(s_t\), estimated noise
power \(\hat{n}_t\), and threshold factor \(\alpha\), a target is detected
when

\begin{equation}
    s_t > \alpha \hat{n}_t .
\end{equation}

For bins without a detected target, the clutter map is updated using an
exponentially weighted moving average:

\begin{equation}
    \hat{n}_{t+1}
    =
    (1-\omega)\hat{n}_{t}+\omega s_t .
\end{equation}

The clutter map is initialized using target-free frames, and the ETCM-CFAR
parameters follow the original method~\cite{cmcfar} without additional
tuning. By avoiding contamination from neighboring body reflections,
ETCM-CFAR reduces energy masking and recovers weak anatomical reflections
that would otherwise be discarded.

\paragraph{Multipath suppression via AoA--AoD consistency.}
Indoor environments introduce dynamic multipath through
person--environment and inter-person reflections. These indirect paths may
have range and Doppler characteristics similar to direct human echoes and
are therefore difficult to remove using static-background subtraction alone.

We suppress multipath-induced ghost points using the geometric consistency
between the Angle of Arrival (AoA) and Angle of Departure
(AoD)~\cite{zhang2026mupose}. For each detected Range--Doppler bin, the
azimuth AoA is estimated from the receiver array, while the azimuth AoD is
estimated from the virtual transmitter array using digital beamforming.

\begin{figure}[t]
    \centering
    \includegraphics[width=0.7\columnwidth]{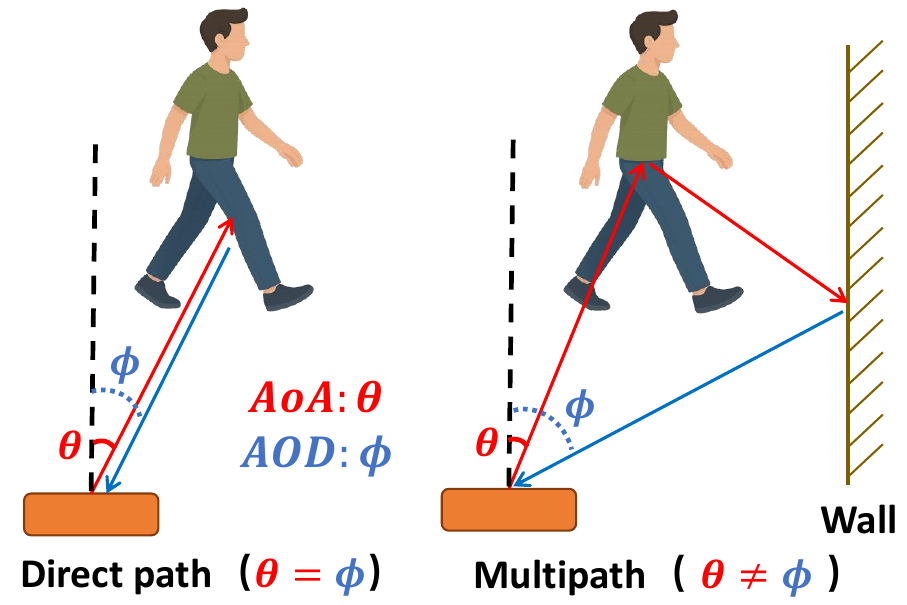}
    \caption{
    Multipath suppression using Angle of Arrival (AoA) and Angle of
    Departure (AoD) consistency. Direct reflections exhibit consistent
    angular estimates, whereas multi-bounce paths introduce angular
    discrepancies.
    }
    \label{fig:AoA_AoD}
\end{figure}

As illustrated in Fig.~\ref{fig:AoA_AoD}, a direct human reflection
preserves a consistent propagation direction between transmission and
reception, resulting in similar AoA and AoD estimates. In contrast,
multi-bounce paths change direction through environmental reflections and
therefore produce inconsistent angular estimates.

We retain detections satisfying

\begin{equation}
    \left|
    \hat{\theta}_{\mathrm{AoA}}
    -
    \hat{\theta}_{\mathrm{AoD}}
    \right| < 2^\circ,
\end{equation}

where the threshold is selected empirically. Only angle-consistent
detections are retained for elevation estimation and Cartesian
reconstruction.

\begin{figure}[t]
    \centering
    \includegraphics[width=\columnwidth]{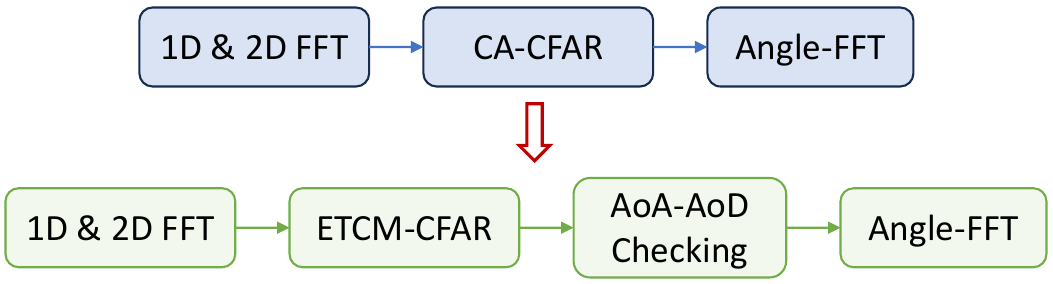}
    \caption{
    Overview of the adopted mmWave point-cloud generation pipeline.
    ETCM-CFAR recovers weak human reflections compared with conventional
    CA-CFAR, while AoA--AoD consistency checking suppresses
    multipath-induced ghost points. Together, they produce denser and
    cleaner point clouds for radar-language modeling.
    }
    \label{fig:point_process}
\end{figure}

As summarized in Fig.~\ref{fig:point_process}, ETCM-CFAR and AoA--AoD
consistency checking improve point-cloud generation from complementary
perspectives. ETCM-CFAR recovers weak body reflections and increases point
density, whereas AoA--AoD consistency checking removes multipath-induced
ghost points. The resulting point clouds provide more complete and reliable
spatial and kinematic evidence for subsequent radar-language modeling.
\section{mmMind-Bench Dataset Details}
\label{sec:supp_data}

\subsection{Data Collection and Sensor Setup}

mmMind-Bench contains 17.9 hours of synchronized multimodal recordings
from 23 participants, including 17 males and 6 females, aged 21--62.
The data were collected in seven real-world indoor environments:
a living room, bedroom, pantry, kitchen, meeting room, office, and open
lounge. These environments exhibit diverse room layouts, furniture
arrangements, sensing distances, and background clutter, providing
substantial variation in both human motion and radar propagation
conditions.

\begin{table}[ht]
\centering
\small
\begin{tabular}{lcc}
\toprule
Parameter & CAL60S244-IB & CAL77S244-IB \\
\midrule
Frequency & 60 GHz & 77 GHz \\
Antenna & 4 Tx $\times$ 4 Rx & 4 Tx $\times$ 4 Rx \\
Frame rate & 15 FPS & 15 FPS \\
Duration & 13.4 h & 4.5 h \\
\bottomrule
\end{tabular}
\caption{
mmWave radar devices used in mmMind-Bench.
}
\label{tab:supp_radar_devices}
\end{table}

\begin{table*}[t]
\centering
\small

\begin{tabular}{p{0.18\linewidth}p{0.76\linewidth}}
\toprule
Category & Behaviors \\
\midrule
Daily activities (11)
&
Walking, sitting, mopping, bending, lying down, using a phone, crouching,
using a laptop, drinking, sweeping, and goose stepping.
\\

Fitness activities (7)
&
Jumping jacks, Tai Chi, squatting, jogging, sit-ups, boxing, and rope
skipping.
\\

Falls (4)
&
Forward fall, backward fall, side fall, and falling against a wall.
\\

Upper-body gestures (5)
&
Left--right motion, push--pull motion, circular motion, up--down motion,
and waving.
\\
\bottomrule
\end{tabular}
\caption{
Behavior categories in mmMind-Bench. Category names are normalized from the
labels used during data collection.
}
\label{tab:supp_action_categories}
\end{table*}

We use two commercial mmWave radar devices from Calterah \cite{calterah}:
the 60-GHz CAL60S244-IB and the 77-GHz CAL77S244-IB. Both devices contain
four transmitting and four receiving antennas and operate at 15 frames per
second. The 60-GHz radar contributes 13.4 hours of recordings and serves as
the primary sensing device because it is designed for indoor sensing
applications. The 77-GHz radar contributes the remaining 4.5 hours,
introducing additional variation in sensing frequency and hardware
characteristics.
Table~\ref{tab:supp_radar_devices}
summarizes the sensing configuration.

An Orbbec Femto Bolt RGB-D camera operating at 30 FPS provides synchronized
color and depth references for dataset annotation and 3D pose acquisition.
The Femto Bolt is compatible with the Azure Kinect software stack, and the
3D skeletons are extracted using Microsoft's official Azure Kinect Body
Tracking SDK rather than a separately trained pose-estimation model. The
RGB-D observations and estimated 3D poses are used only for benchmark
construction, annotation verification, and training-time pose supervision.
At inference time, \systemname{} requires only the mmWave radar point-cloud
sequence and does not use RGB-D observations or pose estimates.

All data collection was conducted under institutional review board approval, and informed consent was obtained from every
participant before recording.

\subsection{Radar Configuration}

The mmWave sensing data are collected using commercial FMCW radars with
four transmitting and four receiving antennas. The dataset includes both
60-GHz and 77-GHz devices, which share the same acquisition configuration.
The radar bandwidth is 3.5 GHz, and each frame contains 512 chirps under
TDM operation. Each transmit antenna contributes 128 chirps, resulting in
an effective Doppler sampling interval of \(320~\mu s\). The raw range
dimension contains 256 FFT bins.

For both radar frequencies, the shared bandwidth and range FFT configuration
provide a range resolution of 5.12 cm and an unambiguous range of 6.51 m.
Although the configured chirp sweep bandwidth is 3.5 GHz, the effective
sampled bandwidth is narrower than the configured bandwidth, resulting in a range resolution of
5.12 cm.
Due to the different carrier wavelengths, the velocity resolution is
0.061 m/s and 0.047 m/s for the 60-GHz and 77-GHz radars, respectively,
with corresponding unambiguous radial velocity ranges of
\(\pm3.87\) m/s and \(\pm3.02\) m/s.

\begin{table}[ht]
\centering
\small
\begin{tabular}{lcc}
\toprule
Parameter & 60 GHz & 77 GHz \\
\midrule
Bandwidth
& \multicolumn{2}{c}{3.5 GHz} \\

Antenna configuration
& \multicolumn{2}{c}{4 Tx $\times$ 4 Rx} \\

Frame rate
& \multicolumn{2}{c}{15 FPS} \\

Chirps per frame
& \multicolumn{2}{c}{512} \\

Chirps per Tx
& \multicolumn{2}{c}{128} \\

Chirp period
& \multicolumn{2}{c}{\(80~\mu s\)} \\

Effective Doppler chirp period
& \multicolumn{2}{c}{\(320~\mu s\)} \\

Range FFT points
& \multicolumn{2}{c}{256} \\

\midrule
Range resolution
& 0.0512 m
& 0.0512 m \\

Unambiguous range
& 6.5059 m
& 6.5059 m \\

Velocity resolution
& 0.0605 m/s
& 0.0471 m/s \\

Unambiguous radial velocity
& \(\pm3.8740\) m/s
& \(\pm3.0190\) m/s \\

\bottomrule
\end{tabular}
\caption{
Radar acquisition and sensing parameters used in mmMind-Bench.
Parameters shared by both radar devices are shown in rows spanning both
columns, while frequency-dependent sensing characteristics are reported
separately.
}
\label{tab:radar_config}
\end{table}

The radar operates at 15 FPS, providing sufficient temporal resolution for
capturing human motion dynamics. The shared acquisition parameters and the sensing characteristics are summarized in Table~\ref{tab:radar_config}.

\subsection{Behavior Categories}

The benchmark covers 27 behavior categories spanning daily activities,
physical exercises, falls, and upper-body gestures. The daily-activity
subset includes both whole-body locomotion and common indoor interactions.
The fitness subset contains periodic and articulated exercise motions. The
fall subset covers different falling directions and interactions with a
wall, while the gesture subset focuses on localized upper-body movements.
The complete action vocabulary is listed in
Table~\ref{tab:supp_action_categories}.

The categories differ substantially in spatial extent and temporal
structure. For example, walking and jogging involve global body translation,
whereas waving and circular motions are dominated by localized limb
movements. Other categories, such as sitting, lying down, and falling,
require recognizing posture transitions and their temporal ordering. This
diversity supports the evaluation of both coarse behavior recognition and
fine-grained kinematic understanding.

\subsection{Temporal Synchronization and Spatial Calibration}

The radar and RGB-D streams are synchronized using acquisition timestamps.
Radar measurements are recorded at 15 FPS, while RGB-D frames are recorded
at 30 FPS. During preprocessing, each radar frame serves as the temporal
reference. The RGB-D frames and 3D pose observations are temporally interpolated
or matched according to the radar timestamps, producing aligned visual
and kinematic references for each radar frame.

The sensing devices are calibrated into a shared room-level world coordinate
system. The RGB-D camera pose is obtained using a checkerboard-based
calibration procedure. Radar position and orientation are calibrated using
the automatic location-attribute calibration approach introduced in
LoCal~\cite{zhang2024local}. The resulting extrinsic transformations map radar
point clouds, RGB-D observations, and 3D poses into the same physical
coordinate system.

In this coordinate system, the \(x\)-\(y\) plane coincides with the room
ground plane, and the \(z\)-axis points vertically upward. Therefore,
\(x\) and \(y\) encode the horizontal location of a radar reflection within
the room, while \(z\) represents its height above the ground. This
world-coordinate representation preserves absolute spatial relations,
supports consistent alignment with 3D poses, and enables questions involving
movement direction, trajectory, and relative position.

\subsection{Clip Construction and Data Splits}

We first partition the continuous recordings into training, validation, and
test sets, and then perform temporal segmentation independently within each
partition. Each recording is segmented into 5-second clips. 
Since the radar operates at 15 FPS, each clip contains 75 radar
frames and approximately 150 synchronized RGB-D frames. Performing the
partitioning before clip segmentation ensures that temporally overlapping
clips never appear across different splits, preventing potential temporal
leakage.

The benchmark adopts participant-, environment-, and session-disjoint
partitions. Specifically, participants, recording environments, and
continuous recording sessions assigned to the validation and test sets are
not included in the training set. This design minimizes leakage from
participant-specific motion patterns, environment-dependent propagation
characteristics, and temporally correlated recordings.

\begin{table}[ht]
\centering
\small
\begin{tabular}{lrrr}
\toprule
Split
& Participants
& Environments
& Duration \\
\midrule
Train
& 17
& 5
& 14.3 h \\

Validation
& 3
& 1
& 1.8 h \\

Test
& 3
& 1
& 1.8 h \\

\midrule
Total
& 23
& 7
& 17.9 h \\
\bottomrule
\end{tabular}
\caption{
Statistics of the mmMind-Bench data partitions.
Partitioning is performed before temporal clip segmentation.
}
\label{tab:supp_split_statistics}
\end{table}

For unseen-action evaluation, complete behavior categories are excluded from
all model-development stages. Specifically, all radar clips and their
associated pose, caption, question-answer, and dialogue annotations from the
held-out categories are removed from radar encoder pretraining,
radar--language alignment, instruction tuning, and hyperparameter
selection. The model must therefore recognize unseen behaviors through
transferable kinematic representations and language knowledge rather than
memorization of previously observed action categories.

\section{Model Architecture and Training Details}
\label{sec:supp_training}

\subsection{Architecture Overview}

Given a 5-second radar clip, the model receives a temporal point-cloud
sequence
\begin{equation}
    \mathbf{X}\in\mathbb{R}^{T\times N\times 4},
\end{equation}
where \(T=75\) is the number of radar frames, \(N=150\) is the maximum
number of retained points per frame, and each point is represented by its
3D position and radial velocity, \((x,y,z,v)\). Frames containing fewer
than 150 detections are zero-padded with a corresponding binary mask,
whereas frames containing more points are randomly subsampled during
training.

The radar encoder first embeds each point into a 256-dimensional feature.
A separate two-layer MLP maps the 3D coordinates to spatial positional
embeddings. Three spatial Transformer layers then model interactions among
points within each frame. A learnable query attends to the point features
and aggregates each frame into a single 256-dimensional token. The resulting
frame-token sequence is processed by eight temporal Transformer layers with
rotary positional embeddings, producing
\begin{equation}
    \mathbf{Y}
    =
    E_{\mathrm{radar}}(\mathbf{X})
    \in
    \mathbb{R}^{75\times 256}.
\end{equation}

All spatial and temporal Transformer blocks use four attention heads and a
1,024-dimensional feed-forward layer. The principal architectural settings
are summarized in Table~\ref{tab:supp_architecture}.

\begin{table}[t]
\centering
\small
\begin{tabular}{lc}
\toprule
Component & Configuration \\
\midrule
Input point feature & \((x,y,z,v)\) \\
Radar frames per clip & 75 \\
Points per frame & 150 \\
Radar token dimension & 256 \\
Spatial Transformer layers & 3 \\
Temporal Transformer layers & 8 \\
Attention heads & 4 \\
Feed-forward dimension & 1,024 \\
Pose joints & 17 \\
Language backbone & Qwen3-8B \\
Language hidden dimension & 4,096 \\
Projector & \(256\!\rightarrow\!4096\!\rightarrow\!4096\) \\
\bottomrule
\end{tabular}
\caption{
Architecture configuration of \systemname{}.
}
\label{tab:supp_architecture}
\end{table}

\subsection{Stage I: Pose-Guided Radar Pretraining}

The first stage trains the radar encoder using synchronized 3D human poses.
A lightweight regression head maps every radar frame token to the 3D
coordinates of \(J=17\) body joints. The pose head consists of LayerNorm,
a 256-dimensional linear layer with GELU activation, and a final linear
layer producing \(17\times3\) coordinates.

The radar encoder is optimized using
\begin{equation}
    \mathcal{L}_{\mathrm{kin}}
    =
    \mathcal{L}_{p}
    +5\mathcal{L}_{r}
    +10\mathcal{L}_{l}
    +2\mathcal{L}_{b},
\end{equation}
where \(\mathcal{L}_{p}\) is the joint-position loss,
\(\mathcal{L}_{r}\) supervises global root motion,
\(\mathcal{L}_{l}\) supervises root-relative joint motion, and
\(\mathcal{L}_{b}\) preserves bone-length consistency. All terms use
\(L_1\)-based supervision except that bone consistency is computed from
the Euclidean lengths of connected joints.

To emphasize articulated motion, the position and local-motion losses use
joint-dependent weights. Torso joints have weight 1, shoulders and hips
have weight 2, elbows and knees have weight 4, and distal joints, including
wrists and ankles, have weight 8. In addition, the final ten frames receive
linearly increasing temporal weights from 1 to 5. This emphasizes the
sequence-end body state, which is particularly important for streaming
behavior analysis.

During radar pretraining, the point clouds and synchronized poses undergo
the same spatial augmentation. Each augmentation is applied independently
with probability 0.5 and includes isotropic scaling in
\([0.8,1.2]\), rotation around the vertical axis in
\([-\pi,\pi]\), horizontal translation within \(5\) cm, and Gaussian
coordinate jitter with standard deviation \(0.01\) m. Padded points remain
masked after augmentation.

We train this stage with AdamW using a batch size of 96, an initial learning
rate of \(6\times10^{-4}\), and weight decay \(10^{-2}\). The first three
epochs use linear warm-up, followed by cosine annealing with warm restarts
(\(T_0=10\), \(T_{\mathrm{mult}}=2\), and
\(\eta_{\min}=10^{-6}\)). Gradients are clipped to a maximum norm of 1.0.
Training uses bfloat16 mixed precision and gradient checkpointing. We set
the maximum number of epochs to 500 and apply early stopping with a patience
of 100 epochs according to validation MPJPE. The checkpoint with the lowest
validation MPJPE is retained. This stage is trained on an NVIDIA RTX 4090.

After pretraining, the pose regression head is discarded. Only the
256-dimensional radar encoder is retained, and no pose input is required
during subsequent training or inference.

\begin{table*}[t]
\centering
\small
\begin{tabular}{p{0.17\textwidth}p{0.25\textwidth}p{0.25\textwidth}p{0.25\textwidth}}
\toprule
Setting
& Stage I: Pose pretraining
& Stage II: Alignment
& Stage III: Instruction tuning \\
\midrule
Trainable modules
& Radar encoder and pose head
& Projector
& Projector and Qwen3-8B LoRA
\\

Supervision
& 17-joint 3D poses
& English/Chinese motion descriptions
& Captions, single-turn QA, and multi-turn dialogues
\\

Batch size
& 96
& 8
& 4
\\

Gradient accumulation
& 1
& 4
& 8
\\

Effective batch size
& 96
& 32
& 32
\\

Learning rate
& \(6\times10^{-4}\)
& \(1\times10^{-3}\)
& \(5\times10^{-5}\) projector;
  \(2\times10^{-5}\) LoRA
\\

Training epochs
& Up to 500 with early stopping
& 8
& 10
\\

Maximum text length
& --
& 256
& 2,048
\\

Checkpoint criterion
& Validation MPJPE
& Validation loss
& Validation loss
\\

Hardware
& NVIDIA RTX 4090
& NVIDIA A800
& NVIDIA A800
\\
\bottomrule
\end{tabular}
\caption{
Training configuration of the three-stage optimization procedure.
The radar encoder is frozen after Stage I.
}
\label{tab:supp_training_stages}
\end{table*}

\subsection{Radar--Language Interface}

The radar encoder produces one token for each of the 75 radar frames. A
two-layer projector maps these tokens from the radar feature space into the
4,096-dimensional embedding space of Qwen3-8B:
\begin{equation}
    \mathbf{R}=P(\mathbf{Y})
    \in\mathbb{R}^{75\times4096}.
\end{equation}
The projector consists of a linear layer, LayerNorm, GELU activation,
a second linear layer, and an output LayerNorm.

To preserve the temporal correspondence between language references and
radar observations, we construct a frame-indexed prefix:
\begin{quote}
\small
\texttt{Frame 0: <|image\_pad|> Frame 1: <|image\_pad|> \ldots
Frame 74: <|image\_pad|>}
\end{quote}
Each placeholder embedding is replaced by the projected radar token from
the corresponding frame. The prefix additionally states that the sequence
contains 75 radar tokens sampled at 15 FPS. Radar-prefix positions are
excluded from the language-modeling loss.

\subsection{Stage II: Radar--Language Alignment}

The second stage aligns the pretrained radar tokens with the language
embedding space. Both the radar encoder and Qwen3-8B are frozen, and only
the projector is optimized. Each radar clip is paired with four description
types: concise English, detailed temporal English, concise Chinese, and
detailed temporal Chinese. The detailed descriptions emphasize posture,
limb movement, trajectory, speed, and temporal evolution.

We use the Qwen chat template and compute the autoregressive cross-entropy
loss only over assistant-response tokens. System prompts, user prompts,
radar-prefix tokens, and padding positions are masked from the loss.

The projector is trained for eight epochs using AdamW with learning rate
\(10^{-3}\), weight decay \(10^{-2}\), and
\(\epsilon=10^{-6}\). The per-device batch size is 8, with four gradient
accumulation steps, resulting in an effective batch size of 32. The maximum
text sequence length is 256 tokens, gradients are clipped to a maximum norm
of 1.0, and all model computation uses bfloat16 precision. Qwen3-8B uses
FlashAttention-2 when available. The best projector is selected according
to validation language-modeling loss. This stage is trained on an NVIDIA
A800.

\subsection{Stage III: Instruction Tuning}

The final stage performs multimodal instruction tuning using captions,
single-turn question answering, and multi-turn dialogues. The radar encoder
remains frozen. The projector is initialized from Stage II and continues to
be optimized, while Qwen3-8B is adapted using LoRA.

LoRA modules are inserted into all attention projections
(\texttt{q\_proj}, \texttt{k\_proj}, \texttt{v\_proj}, and
\texttt{o\_proj}) and feed-forward projections
(\texttt{gate\_proj}, \texttt{up\_proj}, and
\texttt{down\_proj}). We use rank \(r=64\), scaling parameter
\(\alpha=128\), and dropout 0.05. All original Qwen3-8B parameters remain
frozen.

Each clip contains four bilingual captions, 13 single-turn QA pairs,
and four multi-turn dialogues. An annotation is randomly selected from
the corresponding task type during training, while caption prompts and
system instructions are sampled from prompt pools to reduce dependence
on fixed question formulations. 

Single-turn examples supervise only the assistant response. For multi-turn
examples, the full conversation history is supplied as context, while the
loss is computed over every assistant response in the dialogue. User turns,
system instructions, radar-prefix tokens, and padding tokens are excluded
from the loss. Dynamic padding is applied independently within each batch.
The maximum sequence length is 2,048 tokens.

We jointly optimize the projector and LoRA parameters using AdamW with
separate learning rates:
\begin{equation}
    \eta_{\mathrm{proj}}=5\times10^{-5},
    \qquad
    \eta_{\mathrm{LoRA}}=2\times10^{-5}.
\end{equation}
The batch size is 4 with eight gradient accumulation steps, yielding an
effective batch size of 32. Training lasts for ten epochs and uses a linear
learning-rate schedule with a 5\% warm-up ratio. Gradients are clipped to a
maximum norm of 0.5. Training uses bfloat16 precision and
FlashAttention-2 on an NVIDIA A800. The checkpoint with the lowest
validation loss is used for evaluation.

\paragraph{Software and hardware environment.}
Stage-I radar-encoder pretraining was conducted on a cloud server equipped
with an Intel Xeon Gold 6430 CPU, 1.0\,TiB of system memory, and an NVIDIA
GeForce RTX 4090 GPU with 24\,GB of GPU memory. The server ran Ubuntu
22.04.5 LTS with NVIDIA driver 580.76.05. The training environment used
PyTorch 2.6.0, compiled with CUDA 12.4 and cuDNN 9.1.0. Although the
installed NVIDIA driver supports CUDA versions up to 13.0, all experiments
in this environment used the CUDA 12.4 runtime bundled with PyTorch.
Stage-II radar--language alignment and Stage-III instruction tuning were performed on NVIDIA A800
GPUs.

\section{Long-term Behavior Analysis Agent}
\label{sec:supp_long_term_agent}

Although \systemname{} is trained on standardized 5-second clips,
real-world behavior understanding often requires reasoning over longer
recordings. Directly processing long sequences is inefficient because the
number of radar tokens increases with recording duration.

To extend \systemname{} to long-horizon scenarios, we build a hierarchical
behavior analysis agent that combines clip-level radar perception with
temporal memory reasoning. The agent first divides a long radar recording
into timestamped clips and applies \systemname{} to generate concise
radar-grounded descriptions. These descriptions form a temporal memory that
can be organized and queried by a large language model.

Given a user query, the language model first reasons over the temporal memory.
When the stored descriptions are insufficient for fine-grained questions,
the agent retrieves the relevant radar clips and invokes \systemname{} again
with a query-specific instruction. This design separates long-term temporal
reasoning from radar perception: the language model manages temporal
organization and retrieval, while \systemname{} remains responsible for
interpreting radar observations.

\subsection{Temporal Memory Construction}

A long recording is segmented into the same 5-second clips used in
mmMind-Bench. For each clip \(\mathbf{X}^{(i)}\), \systemname{} generates a
radar-grounded caption \(c_i\) describing observable behaviors, posture
changes, motion directions, and temporal transitions.

Each memory entry is represented as

\begin{equation}
m_i =
(
i,
t_i^{\mathrm{start}},
t_i^{\mathrm{end}},
c_i
).
\end{equation}

The complete recording is stored as an ordered temporal memory:

\begin{equation}
\mathcal{M}=\{m_1,m_2,\ldots,m_N\}.
\end{equation}

The temporal memory enables efficient long-range reasoning without requiring
the entire radar sequence to be processed simultaneously.

\subsection{Query-driven Radar Re-perception}

For a given query \(q\), the language model analyzes the temporal memory and
jointly identifies potentially relevant clips and generates a clip-specific
radar instruction for each retrieved clip:

\begin{equation}
\pi(q,\mathcal{M})
=
\bigl\{(i,q_i)\mid i\in\mathcal{K}\bigr\},
\end{equation}
where \(\mathcal{K}\) is the set of retrieved clip indices, \(q_i\) is the
radar instruction for clip \(i\) generated from the original query and its
temporal context, and \(\pi\) denotes the query-planning process.

For questions requiring detailed motion evidence, the selected clips are
re-analyzed by \systemname{}:

\begin{equation}
r_i=
\operatorname{mmMind}(\mathbf{X}^{(i)},q_i),
\quad i\in\mathcal{K}.
\end{equation}

The final response integrates temporal context from memory with detailed
radar perception results. This retrieval-based design allows the system to
scale to long recordings while allocating radar computation only to
query-relevant intervals.

\bibliography{mmMind}

\end{document}